\documentclass[compsoc]{IEEEtran}

\usepackage{verbatimbox}

\usepackage{cite}
\usepackage[linesnumbered,ruled,vlined]{algorithm2e}

\usepackage{xcolor}

\usepackage{amsmath,amsthm,amssymb}

\usepackage{graphicx}
\usepackage{multirow}

\usepackage{adjustbox}

\usepackage{url}

\DeclareGraphicsExtensions{.pdf,.jpeg,.png}

\begin{document}

\title{A chaotic maps-based privacy-preserving distributed deep learning for incomplete and Non-IID datasets}

\author{
\IEEEauthorblockN{Irina Ar\'evalo}\\
\IEEEauthorblockA{Universidad Pablo de Olavide, Sevilla (Spain)\\
Email: iarebar@alu.upo.es}\\[0.9cm]
\and
\IEEEauthorblockN{Jose L. Salmeron}\\
\IEEEauthorblockA{CUNEF Universidad, Madrid (Spain)\\}
\IEEEauthorblockA{Universidad Aut\'onoma de Chile (Chile)\\
Email: joseluis.salmeron@cunef.edu}}

\maketitle

\begin{abstract}
Federated Learning is a machine learning approach that enables the training of a deep learning model among several participants with sensitive data that wish to share their own knowledge without compromising the privacy of their data. In this research, the authors employ a secured Federated Learning method with an additional layer of privacy and proposes a method for addressing the non-IID challenge. Moreover, differential privacy is compared with chaotic-based encryption as layer of privacy.
The experimental approach assesses the performance of the federated deep learning model with differential privacy using both IID and non-IID data. In each experiment, the Federated Learning process improves the average performance metrics of the deep neural network, even in the case of non-IID data.
\end{abstract}


\IEEEpeerreviewmaketitle

\section{Introduction}



Privacy-preserving\footnote{Preprint - Published in IEEE Transactions on Emerging Topics in Computing} machine learning models are designed to protect the privacy of individuals whose data is used to train the model. This can be achieved through various techniques, such as using Federated Learning (FL) to train a model on decentralized datasets without sharing the raw data, or using secure multiparty computation to allow multiple parties to collaboratively train a model without revealing their individual data. Privacy-preserving models are par\-ti\-cu\-larly important in sensitive applications, such as healthcare or finance, where protecting the personal information of individuals is a top priority. 
Nevertheless, such additional privacy layer could lower the model performance both in accuracy and time. 

In this research, the authors compare the results of an unsecured federated model and a secured model built using differential privacy and chaotic maps as its encrypting layer. In each experiment this research proves that the federation process improves the averaged performance metrics of a deep neural network for the participants, with disregard of whether the data has been evenly split among them or there are differences between the amount of data each participant has, and that the performance with or without the privacy layer are similar, meaning that the additional security does not worsen the model's results thanks to the federation process. 

Moreover, to ensure the simulation closely resembles a real-world implementation of Federated Learning, the authors have taken into account a scenario where one of the participants or clinical centers possesses incomplete data with a distinct structure. This situation may arise, for example, when a variable is unavailable in the dataset of one participant. In such cases, this proposal enables other participants to privately share the distribution of the missing variable, allowing for the imputation of the missing data in the dataset of the participant who lacks that variable. The rational behind this proposal is that the federation process and its multiple iterations will average the model performance even in the case when one of the participant's features has been imputed. 

The main contributions of this paper are two fold:
\begin{itemize}
    \item This proposed FL extension aims to handle datasets that are incomplete or contain missing values, as well as datasets that are non-IID (non-Independently and Identically Distributed). In certain scenarios, the data used in the FL process may have missing values, lack completeness or uneven distribution across the participating devices or nodes.
    \item An efficient and secure method for encrypting distributed mo\-dels based on chaotic maps. Chaotic maps possess inherent complexity and unpredic\-ta\-bi\-lity, which makes them resistant to conventional cryptographic attacks. Furthermore, their non-linear nature enhances their security. Additionally, the deterministic characteristics of chaotic maps make them an efficient encryption method.
\end{itemize}

The proposed extension aims to address these issues and enable the use of FL on several categories of non-IID datasets. As best of our knowledge, this approach is a novelty. The specific details of the extension, as well as its performance and effectiveness, are described in the paper. The categories of non-IID data analysed are the following ones:
\begin{itemize}
    \item Partial overlapping attribute skew
    \item Full overlapping attribute skew
    \item Label distribution skew
    \item Attribute and label skew
    \item Quantity skew.
\end{itemize}

The rest of this paper is organized as follows. We discuss the theoretical background and related work in section \ref{related.work}. The methodological proposal is described in section \ref{methodological.context}.
Section \ref{experiments} outlines the details of the experimental approach and the results. Finally, the authors draw a conclusion in section \ref{conclusions}.


\section{Related work}\label{related.work}
Recently, deep learning has achieved remarkable results in different domains, such as object classification \cite{esteva.2017} and self-driving cars \cite{ramos.2017}. Deep learning is a subset of machine learning algorithms that models high level abstraction using computational architectures that allow non linear transformations in the data in the form of a neural network \cite{guerrero.2020}. 

A deep neural network for a supervised problem learns from processing many labeled examples through its layers \cite{GoodBengCour16, lecun15}. Layers are composed of a number of interconnected nodes which contain an activation function that polarizes network activity. This function includes nonlinear behaviour and helps it to become steady. A common activation funcion is ReLU (Rectified Linear Unit), which both the function and its derivative are monotonic. The function returns 0 if it receives any negative input, but for any positive value x, it returns that value back, and thus it gives an output that has a range from 0 to infinity. 

The labeled examples to train the network are supplied via the input layer, which communicates to one or more hidden layers where the actual processing is done via a system of weighted connections. The hidden layers then link to the output layer. The training of the neural network is done via backpropagation, an algorithm that modifies the weights of the network by computing the gradient of the loss function with respect to those weights for a single input–output example.

\subsection{Non-IID data}

The data used to train a model on each client in Federated Learning often depends on the usage patterns of specific local devices, resulting in data distributions among connected clients that can vary significantly from one another. This phenomenon is known as Non-Independent and Identically Distributed (Non-IID) data \cite{mcmahan.2017}. Zhe et al. \cite{zhu.2021} propose several categories of non-IID data:
\begin{itemize}
    \item \underline{Attribute skew}. This category includes several subcategories: \textit{Non-overlapping attribute skew}: It means that data attributes across the clients are mutually exclusive. \textit{Partial overlapping attribute skew}: In this case, some portions of the data attributes can be shared with each other. \textit{Full overlapping attribute skew}: Data attributes are the same in all participants but the attributes distributions can be different.
    \item \underline{Label skew}. This category includes several subcategories: \textit{Label distribution skew}: Label distributions on the clients are different. \textit{Label preference skew}: The label distribution is different on the client data, although the distribution of the attributes is the same. \textit{Temporal skew}: The focus is on addressing distribution skewness in temporal data, which encompasses spatio-temporal data as well as time series data.
    \item \underline{Attribute and label skew}. Different clients hold data with different labels and different attributes.
    \item \underline{Quantity skew}. The number of training data varies across different clients.
\end{itemize}

In general, non-IID datasets can be challenging to analyse because they often contain a high degree of variance and may not be representative of the overall population. This paper is focused on the performance analysis of differential privacy on FL in the IID data and several flavours of non-IID data. 
This is a challenging issue because the features of connected clients are different from each other \cite{zhu.2021}. 

\subsection{Federated Learning}
\label{fl}

Federated Learning, proposed by McMahan et al. \cite{mcmahan.2016} and further developed in Konecny et al. \cite{konecn.2016} and McMahan and Ramage \cite{mcmahan_ramage.2017}, is a distributed Machine Learning approach in which the participants collaborate to train a model with their private data by updating that model in their infrastructure and then sending the parameters to an aggregation node. The participants own the data and train the partial models. The aggregation node then federates the participant's models to obtain a global model trained with private data. This method can be iterated as many times as desired. 

The use of FL with non-IID datasets has been studied in the literature  \cite{zhu.2021}. For instance Zhao et al. \cite{zhao.2018} train convolutional neural networks (CNNs) on MNIST, CIFAR-10 and Speech commands datasets and find the reduction in the test accuracy of the federated averaging for non-IID data. Also, Wang et al. \cite{wang.2020b} optimise Federated Learning on Non-IID Data with Reinforcement Learning. Chen et al. \cite{chen.2020} proposed an asynchronous online FL framework, where the edge devices perform online learning with continuous streaming local non-IID data and a central server aggregates model parameters from clients. None of the previously mentioned research focus the analysis of FL's performance with non-IID data and Differential Privacy or Chaotic Maps, as proposed in this paper.

\subsubsection{Federated Learning architectures}

There are two main Federated Learning architectures \cite{yang.2019}:
\begin{enumerate}
    \item Coordinated or centralised (client–server): It consists of a central server, that delivers the model architecture, performs the aggregation tasks, manages the communications, and delivers the model architectured , and a set of data silos or participants.
    \item Swarm Learning (Peer-2-Peer): This architecture does not need any central server because all the nodes play the role simultaneously of central server and data silos. In this architecture, the Federated Learning process is triggered by one of the nodes.
\end{enumerate}

The main advantage of Federated Learning is the training of a model in the private data of several participants that wish to maintain avoid data-sharing while improving their models \cite{duan.2022}. This approach allows the use of heterogeneous data among the participants. It also allows the use of more accurate models with low latency, ensuring privacy and less power consumption. The process of a coordinated Federated Learning process is as follows:

\begin{enumerate}
\item The central server sends a model to each participant. If this is the initial iteration the federated model is proposed by the central server.
\item Each participant trains the received model using their own private data.
\item Each participant sends the parameters of the model in a private way (usually encrypting the data to be sent, see next subsection) to the central server.
\item The central server aggregates the partial models through their parameters and builds the federated model. 
\item The central server checks a termination condition in either accuracy of the model in a test dataset or number of iterations. If it is accomplished, the FL process ends, otherwise we iterate from step 1.
\end{enumerate}

The development of a Peer-2-Peer Federated Learning process is similar with one of the nodes taking the role of the central server. In any case, the target of the federated model is to minimize the total loss for all participants, computed as follows:

\begin{equation}
     \mathcal{L}^* = \frac{1}{n} \sum_{i=1}^n \mathcal{L}(\mathcal{D}_i,\Phi)
\end{equation}

\noindent where $n$ is the number of participants, $\Phi$ is the federated model parameters, $\mathcal{D}_i$ is the dataset of the participant $i$, $\mathcal{L}^*$ is the loss function for the federated model, and $\mathcal{L}_i(\cdot)$ is the loss function for each participant in the federation. 


\subsubsection{Federated Learning categories}
Regarding the nature of the data, Federated Learning can be categorised \cite{li.2020,yang.2019} into three sets (Horizontal Federated Learning, Vertical Federated Learning and Federated Transfer Learning).

In Horizontal Federated Learning, the features space is overlapped across data silos, but the samples space is different in data locations. This approach is the original Federated Learning proposal but it still presents challenges. For instance, an innovative approach named hierarchical heterogeneous horizontal Federated Learning faces limited labeled entities in Horizontal Federated Learning \cite{li.2020}. In this research, the lack of labeled instances is mitigated by adapting the heterogeneous domain multiple times by using each participant as the target domain each time. 

Vertical Federated Learning is needed when the features space has a partial or low overlap across data silos, but the samples space is nearly the same across those data locations. Unlike the case of horizontal Federated Learning, the aggregation of the entire data set in a single data silo to train a global model would not work in vertical Federated Learning. Some vertical proposals have been developed in \cite{cheng.2019,lee.2020}.

Moreover, the data does not share a sample space or a feature space in most cases. Federated Transfer Learning approach proposed by \cite{Liu.2020} generalise Federated Learning when it comes to common parties with small intersection. This proposal can be easily adapted to various secure, Machine Learning endeavours with minimal modification to the existing model and provides the same level of accuracy as non-privacy-preserving transfer learning.

\subsubsection{Federated Learning challenges}

There are several challenges that must be addressed in Federated Learning in order to effectively protect the privacy of enterprises and users \cite{zhang.2021}. These include:

\begin{enumerate}
\item \textit{Ensuring privacy protection}: Federated Learning was designed to protect the privacy of data in machine learning, and it is important to ensure that the training model does not reveal users' private information or that the model itself is altered.
\item \textit{Overcoming the lack of sufficient data}: In conventional machine learning, a large amount of data is typically needed to train a model with optimum performance. However, in a distributed environment, the amount of data on each device may be insufficient and collecting all the data in a centralized manner can be too expensive or legally prohibited. Federated Learning allows each device to use its own local data to train a local model, which is then aggregated with the models of other devices to create a global model.
\item \textit{Dealing with statistical heterogeneity}: There are many edge devices in the federated environment, and the data held by these devices may not be evenly distributed or similar in structure (i.e., it may exhibit skew). 
\end{enumerate}

This paper addresses all of these challenges simultaneously. The authors test the use of differential privacy and chaotic maps for improving privacy protection, conduct experiments with participants that have very limited amounts of data, and examine the impact of different skew and overlap of attributes among participants.

\section{Privacy-preserving techniques}
In this paper the authors are testing two privacy-preserving techniques: differential privacy and chaotic maps-based encryption.

\subsection{Differential Privacy}

The Federated Learning process guarantees that the sharing of private data is not needed to train the federated model. However, there are still risks associated with the transmission of such information, like model poisoning, potential attacks to reconstruct the model or the training data from the parameters that the participants send to the central server, or the use of attack models \cite{pmlr-v108-bagdasaryan20a, Wang19}. Therefore, there have been several advances in the use of privacy-preserving methods \cite{abadiDP16, Acar17, Kaissis2020SecurePA, Hu20, Cheng20} in Federated Learning. 

In this research the authors have applied differential privacy to ensure the security of the system. Differential privacy is a widely-used standard for privacy guarantee of algorithms operating on aggregated data. In general, a randomised algorithm $A(D)$ satisfies $\varepsilon, \delta$-differential privacy if for all
datasets $D$ and $D'$ that differ in a single record, and for all sets $S \in R$, where $R$ is the range of $A$, 
\begin{equation}
P(A(D)\in S) \leq \textrm{exp}(\varepsilon) P(A(D')\in S) + \delta
\end{equation}
\noindent where the probability $P$ is taken over the coin tosses of $A$ and $\varepsilon$ and $\delta$ are non-negative numbers. This means that no single record in the dataset has a significant impact on the output of the algorithm. 

The authors add a differential privacy layer to a deep network using the Differentially Private Stochastic Gradient Descent algorithm 
that modi\-fies the optimization process in a deep network adding some noise \cite{abadiDP16}. 
This algorithm trains the model by obtaining the parameters $\theta$ via minimizing the empirical loss function $\mathcal{L}.$ 

Here we assume that the gradient of the loss function has a bounded $L^2$ norm, therefore we ask for the loss function to be a Sobolev function, $\mathcal{L}\in \mathcal{W}^{1,2},$ which is a weaker condition than being a Lipschitz function. Nevertheless, in order to ensure the convergence of the algorithm with non-IID data, we will additionally ask for the gradient of the loss, $\nabla\mathcal{L},$ to be a Lipschitz function. The inputs of the algorithm are the examples $\{x_i\}_{i=1}^N$, the loss function $\mathcal{L}(\theta)$, the learning rate $\eta_t$, the noise scale $\sigma$, the group size $L$, and the Sobolev norm of the loss function $C$.

%

This additional privacy layer is expected to lower the model performance, both in accuracy and in the training time, due to the extra computations and the necessity of finding the privacy cost $\varepsilon, \delta.$

The use of differential privacy in FL has already been studied in works such as \cite{wei.2020}, where they compare the accuracy for an MLP trained on MNIST data for Different Privacy values $\varepsilon,$ number of participants and iterations to experimentally evaluate their algorithm. The application of differential privacy in FL with non-IID dataset is not a novelty either. Zhao et al. \cite{zhao.2018} have previously applied differential privacy to non-IID datasets, including in cases where participants only received data from a single class. In contrast, this paper addresses the challenge of missing features, rather than just a single class of the target, in the context of incomplete and non-IID datasets. 

\subsection{Chaotic maps-based privacy-preserving}\label{chaotic}
Chaotic maps are a branch of mathematics that investigates dynamic systems capable of generating highly randomized states. These states exhibit complete disorder and apparent irregularity, yet their evolution is determined by the initial conditions of the system. Chaotic maps exhibit sensitive dependence on initial conditions and generate complex, unpredictable behavior. This unpredictability can be harnessed for encryption purposes.

Chaotic maps algorithms for encryption are highly regarded for their ability to deliver a combination of high speed, reasonable computation, and strong security. It's worth noting that the specific implementation and design choices for encryption with chaotic maps can vary. Different chaotic maps can be used, such as the logistic map, Henon map, or Lorenz system, depending on the desired properties and security requirements. In this research, logistic map is the selected map for testing our proposal.

The logistic map, a recurrence relation of degree 2 or polynomial mapping, is widely recognized as an archetypal instance that demonstrates the emergence of complex and chaotic behavior from simple nonlinear dynamical equations. The logistic map is defined as
\begin{equation}
    x_{i+1}= r\cdot x_i\cdot(1-x_i)
\end{equation}
\noindent where the parameter $r$ fall within the interval $[0, 4]$ in order to ensure that $x_n$ remains bounded on $[0, 1]$. When $r\in [3.57, 4]$ the logistic map is chaotic \cite{ibitoye.2023}. In this research, the value of $r$ is assigned as $3.8$ to ensure chaotic behaviour.

\begin{algorithm}
\DontPrintSemicolon
\SetAlgoLined
\KwData{Original plain data ($\mathcal{D}$)}
\KwResult{Cipher data ($\Gamma$)}

\textbf{Key Generation:} Generate the $r$ parameter\;
\textbf{Initialization:} Choose an initial value $x_0$\;
\textbf{Encryption:} \\
\For{$i=1$ \KwTo $n$}{
    \tcc*[l]{Calculate the chaotic value };
    $x_{i+1} = r \cdot x_i \cdot (1 - x_{i-1})$\;
    $\Gamma[i] = \mathcal{D}[i] \oplus \textrm{Frac}(X[i])$\;
}
\textbf{Output:} The cipher data obtained from the encryption process\;
\caption{Logistic map-based encryption}\label{alg:enc}
\end{algorithm}

\begin{algorithm}
\DontPrintSemicolon
\SetAlgoLined
\KwData{Cipher data ($\Gamma$)}
\KwResult{Original plain data ($\mathcal{D}$)}

\textbf{Key Generation:} Generate the $r$ parameter\;
\textbf{Initialization:} Choose an initial value $x_0$\;

\textbf{Decryption:} \\
\For{$i=1$ \KwTo $n$}{
    \tcc*[l]{Calculate the chaotic value};
    $x_{i+1} = r \cdot x_i \cdot (1 - x_{i-1})$\;
    $\textrm{Frac}[i] = \mathcal{D}[i] \oplus \Gamma(X[i])$\;
}
\textbf{Output:} The plain data obtained from the decryption process\;
\caption{Logistic map-based decryption}\label{alg:dec}
\end{algorithm}

Encryption with chaotic maps is a method of encrypting data using chaotic dynamics (Algorithm \ref{alg:enc}). In the case of the encryption algorithm with the logistic map, XOR  ($\oplus$) is applied between each element of the data set (whether it is plain data or cipher data) and the fractional part of the chaotic value generated by the logistic map at that moment. XOR the $i$-th element of the plain data with the fractional part of $x_i$ to obtain the $i$-th element of the cipher data. It is important to note that XOR ($\oplus$) is a bitwise operation, which means that it is applied independently to each corresponding pair of bits in the data elements and chaotic values. This allows for a reversible operation (Algorithm \ref{alg:dec}), as performing XOR between the encrypted data and the same encryption key (or parameter) will yield the original data.

The distinct characteristics exhibited by chaotic systems, including determinism, ergodicity, and sensitivity to initial conditions, make them a compelling option for constructing cryptographic systems. These properties share similarities with the desirable properties of a robust cryptosystem, such as confusion and diffusion. One of the advantadges of Chaos-based encryption techniques is their computational efficiency \cite{zia.2022}. The encryption process with chaotic maps is as follows:

\begin{enumerate}
    \item Key Generation: Chaotic maps require a secret key to initialize the map. The key should be kept secret, as it determines the encryption/decryption process. In the case of the logistic map, the parameter $r$ determines the chaotic behavior of the map.
    \item Chaotic Map Iteration: The chaotic map takes a value and generates iteratively a new value based on its mathematical definition and the previous value. The inherent chaotic nature of the map guarantees that even a slight alteration in the initial value can yield a significatively different output.
    \item Offuscation: The chaotic map's output could be combined with the original data through offuscation operations. The aim of this stage is to make the relationship between the original and the encrypted data as complex and nonlinear as possible.
    \item Iterations and key updating: During the encryption process, it is common to employ multiple iterations of the chaotic map along with key updates. Following each iteration, the key may undergo changes to introduce additional randomness and strengthen the security of the encryption..
    \item Output: The final output of the encryption process is the cipher data, which is the encrypted form of the original plain data. It should appear random and be statistically independent of the original data.
    \item Decryption: The same chaotic map is applied iteratively to the cipher data using identical initial conditions, parameters and key as in the encryption phase to retrieve the original plain data.
\end{enumerate}

Encryption with chaotic maps offers certain advantages, such as a high degree of randomness, sensitivity to initial conditions, and resistance to various attacks. However, it also poses challenges in terms of stability, security analysis, and the need for efficient chaotic map implementations. it's important to note that in the context of chaotic maps, the terms \textit{encryption} and \textit{decryption} might not be the most accurate. Chaotic maps are primarily used for generating pseudorandom sequences or for generating chaotic beha\-viour, rather than encryption and decryption.

\section{Methodological proposal}\label{methodological.context}

The extension of Federated Learning proposed to combine the use of an additional encryption layer with partial and full overlapping attribute, differents skews and non-IID datasets is as follows: 
\begin{enumerate}
    \item As a first step, a central server will send an untrained deep learning model to the participants.
    \item If one of them does not have a complete dataset, meaning that one of the features is missing (and therefore the features are non-IID), the server will also send, in a encrypted fashion, the distribution of the feature for any other participant so that the lacking feature can be imputed.
    \item Then all the participants will split their data into train/test/validation datasets and train the model in their training data. Then, they will send the parameters of the model back to the central server, using one of the three possible different approaches to this step: 
    \begin{enumerate}
        \item the first one where the data is sent without any additional security measures, 
        \item the second where the data is encrypted using either differential privacy to avoid privacy issues, in order to compare if the use of this privacy-preserving layer affects the results of the model, 
        \item and the third, where the data is encrypted using a chaotic map and the process ends with the decryption of the offuscated data, as described in subsection \ref{chaotic}.
    \end{enumerate}
    \item The server then aggregates the parameters of the local models to find a global model, and iterates this process to improve the global accuracy. The most common aggregation method is Federated Averaging, which is just a weighted average of the network weights across training sites. 
\end{enumerate}

The federated model is evaluated in each participant's test data. The convergence of the differential privacy process is guaranteed by the work in \cite{wei.2020}, where the authors apply Differential Privacy to Federated Learning.

The convergence of the chaotic map encryption process is ensured by the encryption-decryption process as described in subsection \ref{chaotic}.

\section{Experimental approach}\label{experiments}



The scenario for the experiments is as follows. The authors are assuming there are several participant facilities that have private health data from their patients. This information is used to train a deep learning-based classification model to diagnose some specific disease with high security standards.

Nevertheless, the amount of available data is inadequate in some cases for training a robust model. One naive solution is to share their data with others participant facilities or an intermediary to train a model with all their shared data. But since the participants are dealing with extremely sensitive information, it is unlikely they would accept that solution, and it even raises the question of whether it is compliant with data regulations. 
Moreover, there exists the possibility that the distribution of data is not uniform among all the participants. In this simulation, the authors consider the case where one of the participants does not have one of the features the others have. 

%
The selected initial model is a dense neural network, made of five dense layers, each of them followed by a ReLU function and a dropout layer of 0.3 for regularization, with a learning rate  0.01 and the loss function is binary cross-entropy. 

%
%
%
For the experiments we will assume there are five different participants. In a first test all of them will have the same amount of data, obtained from a evenly split dataset, but we will also consider the case where each participant has a different dataset size.

In particular, the authors have performed three additional experiments with uneven splits: the first one will be a random split among the participants, but in the remaining two we have forced that there are several participants with a very small number of samples (less than 10\% of the samples). The amount of positive cases will also vary from one participant to another. In any case, each participant's data will be split into a train and a test dataset. As it is customary, the models will be trained in each participant's train data, and the evaluation metrics will be obtained from each participant's test data. The final performance metrics will be averaged. 

We also simulate the case where one of the participant does not have data on one of the variables. In that case, we use the encrypted sharing of data to send the distribution of that variable in one of the other participants in a private way, and then proceed to impute the mean of the value. The distribution of the missing feature is computed in another participant's data and is sent via the encryption used for sharing the model data to the participant without the feature. Then, the authors apply the $L2$ imputation using the features' distribution.

For each experiment the authors will perform two different FL approaches, as detailed in Section \ref{methodological.context}: one with a layer of differential privacy, and another one without it. The differences in results between the two types of experiments are compared to understand whether the encrypted version offers similar results. To compare the performance of different models and methods, the authors applied metrics (accuracy and F1 score) in a set of tests for each participant.

The results are presented in tables with the metrics for each participant, and the columns are as follows: the size or percentage of the original dataset that each participant has the percentage of positives in that participant's dataset, and  the accuracy and the F1 score on a test set before and after the Federated Learning process, for the non-private, the differentially private, and the chaotic map-based privacy approaches. For each experiment there will be two tables, one for the case when there are no missing features, and one for the case when the fifth participant does have a missing column. 

\subsection{Experiment 1}

The dataset for the first experiment is the Breast Cancer Wisconsin. It was created by the University of Wisconsin Hospital at Madison, and is publicly available \cite{dua.2019}. 
More details can be found in \cite{street.1993}, \cite{mangasarian.1995}.

The results of the Federated Learning process for this dataset are shown in Tables \ref{results.breast} and \ref{results.breast.imp}. As previously mentioned, the authors have tested two different scenarios:
\begin{itemize}
    \item All of the participants have data with a consistent structure. 
    \item One participant is missing one of the features in their data.
\end{itemize}

The authors have also considered four different data partitions: the first one is an evenly split dataset for every Participant, were each of them has a 20\% of the Brest Cancer dataset, while the remaining three tests include uneven sets, the first one a random partition and the remaining two with sharp differences where several participants have very small datasets, like 2\% and 6\% in the case of the third test and 5\% and 5\% in the case of the fourth test. There also are an uneven percentage of positives in these splits, including some participants with more than a 65\% of positive samples and others with less than a 25\%. With this setting, it is possible to test some hypothetical cases where a group of participants want to share secure information and a private model even in the case where one or more of the participants have much less information to share than the rest. 

As the results show, the Federated Learning process improves the averaged performance metrics for all participants in every case, both the even and the uneven partitions, and the private and non-private approaches. In particular, when there is no missing data, the Federated Learning process improves the metrics for the experiments without an additional privacy layer from 0.9735 to 0.9826 in average accuracy for all participants. In average F1 the score goes from 0.9538 to 0.9667 for all participants for the evenly split case. In average accuracy from 0.9609 to 0.9672, and from 0.9259 to 0.9333 in average F1 in the first case of uneven splits. From 0.9097 to 0.9536 in average accuracy and from 0.8946 to 0.9397 in average F1 in the second case of uneven splits. From 0.9679 to 0.9778 in average accuracy and from 0.9533 to 0.9778 in average F1 for the last uneven split. 
This improvement is to be expected due to the convergence of the federation process and the use of iterative local models. 

In the case of the experiments with an additional differential privacy layer added, the average accuracy increases from 0.9470 to 0.9648 while the average F1 goes from 0.9051 to 0.9444 with even splits. The average accuracy goes from 0.9424 to 0.9593 and the average F1 from 0.9180 to 0.9451 for the first uneven split. For the second uneven split the average accuracy increases from 0.9034 to 0.9401 and the average F1 from 0.8513 to 0.8897. In the last uneven split the average accuracy goes from 0.9230 to 0.9848 and the average F1 from 0.8287 to 0.9287. 

Moreover, the results for the private approach, where differential privacy is used, are in general terms very similar to the non-private approach, with a small decrease in accuracy and F1 in the case of the balanced datasets, and both decreases and increases in the performance metrics in the imbalanced examples. 
This outcome is more surprising, since given the additional privacy layer used in the models one could expect worse accuracy metrics. Nevertheless, the federation process, iterating the averaging of all the local models, is able to maintain the performance metrics of the non-differentially private model.

The results for the experiments with missing data are also very similar to the previous outcomes, with a small decrease in the performance metrics with respect with non-missing data, but still achieving high performance metrics. Once again this result is surprising, not because of the lower performance metrics, which is understandable given the missing data, but for its limited nature, since one could anticipate a larger decrease. Nevertheless, the iterative federation process averages the metrics and reduces its diminution after several repetitions.

More explicitly, the average accuracy for the experiments without an additional privacy layer increase from 0.9391 to 0.9826 in the case of even splits. In the first uneven split from 0.9321 to 0.9397. From 0.9345 to 0.9470 in the second, and from 0.8961 to 0.9815 in the last one. The average F1 score goes from 0.9136 to 0.9825 in the even split, from 0.9016 to 0.9407 in the first uneven split, from 0.9286 to 0.9331 in the second, and from 0.7766 to 0.9698 in the third. 

In the case of the additional differential privacy layer, the model improves the average accuracy from 0.9478 to 0.9652 in the evenly split datasets, from 0.9353 to 0.9536 in the first uneven split datasets. From 0.9157 to 0.9913 in the second split, and from 0.9294 to 0.9657 in the last one, and the increment of the average F1 score goes from 0.9339 to 0.9697 in the split with even data for every participant. From 0.9213 to 0.9365 in the first split with uneven data, from 0.9123 to 0.9818 in the second, and from 0.9095 to 0.9409 in the last one. 

Finally, in the case of the chaotic map approach with no missing data, for the even split datasets the accuracy improves from 0.9561 to 0.9652 while the F1 score goes from 0.9331 to 0.9381, in the first uneven split the accuracy increases from 0.9424 to 0.9778 and the F1 from 0.9228 to 0.9750, in the second uneven split the accuracy goes from 0.9301 to 0.9679 and the F1 from 0.9190 to 0.9618, and in the third uneven split the metrics improve from 0.9571 to 0.9655 in the case of the accuracy and from 0.9244 to 0.9655 in the case of the F1 score.

When dealing with missing data in the chaotic map experiment, the accuracy increases from 0.9301 to 0.9478 when the data is split evenly, from 0.9614 to 0.9637 with the first uneven split, from 0.9413 to 0.9736 in the second uneven split, and from 0.9773 to 0.9909 in the third uneven split, whereas the F1 score goes from 0.9123 to 0.9228, from 0.9252 to 0.9267, from 0.9157 to 0.9554, and from 0.9706 to 0.9913 in the even split and first, second and third uneven split respectively.

\begin{table*}[ht]
\centering
\tiny
\caption{Experiment 1 (no missing data)}\label{results.breast}
\begin{tabular}{|c| c| c|| c| c|| c| c|| c| c|| c| c|| c| c|| c| c|}
\hline
 \textbf{Par-}& & & \textbf{Accuracy} &         \textbf{Accuracy} & \textbf{F1} & \textbf{F1} & \textbf{Accuracy} &         \textbf{Accuracy} & \textbf{F1} & \textbf{F1} & \textbf{Accuracy} &         \textbf{Accuracy} & \textbf{F1} & \textbf{F1}  \\
 \textbf{tici-} & \textbf{Size} & \textbf{pos (\%)} & \textbf{pre-FL} &         \textbf{post-FL} & \textbf{pre-FL} & \textbf{post-FL} & \textbf{pre Pri-} &         \textbf{post Pri-} & \textbf{pre Pri-} & \textbf{post Pri-} & \textbf{pre encr.} &         \textbf{post encr.} & \textbf{pre encr.} & \textbf{post encr.}  \\
 \textbf{pant} & & & & & & & \textbf{vate FL} & \textbf{vate FL} & \textbf{vate FL} & \textbf{vate FL}  & \textbf{FL} & \textbf{FL} & \textbf{FL} & \textbf{FL} \\
\hline
1 & 20\% & 47\% & 0.9546 & 1.0000 & 0.9231 & 1.0000 &  0.9091 & 0.9545 & 0.8750 & 0.9231 & 0.9545 & 1.0000 & 0.9333 & 1.0000\\
2 & 20\% & 41\% & 1.0000 & 1.0000 & 1.0000 & 1.0000 &  1.0000 & 0.9565 & 1.0000 & 0.9655 & 0.9565 & 1.0000 & 0.9655 & 1.0000\\
3 & 20\% & 31\% & 0.9565 & 0.9130 & 0.9231 & 0.8333 &  0.8696 & 0.9130 & 0.7273 & 0.8333 & 0.9130 & 0.9130 & 0.8333 & 0.8333\\
4 & 20\% & 35\% & 1.0000 & 1.0000 & 1.0000 & 1.0000 &  0.9565 & 1.0000 & 0.9231 & 1.0000 & 1.0000 & 1.0000 & 1.0000 & 1.0000\\
5 & 20\% & 32\% & 0.9565 & 1.0000 & 0.9231 & 1.0000 &  1.0000 & 1.0000 & 1.0000 & 1.0000 & 0.9565 & 0.9130 & 0.9333 & 0.8571\\\hline
Avg & - & - & 0.9735 & \textbf{0.9826} & 0.9538 & \textbf{0.9667} & 0.9470 & \textbf{0.9648} & 0.9051 & \textbf{0.9444} & 0.9561 & \textbf{0.9652} & 0.9331 & \textbf{0.9381}\\
\hline\hline
1 & 19\% & 28\% & 1.0000 & 1.0000 & 1.0000 & 1.0000 & 0.9000 & 0.9500 & 0.8000 & 0.8889 & 1.0000 & 1.0000 & 1.0000 & 1.0000\\
2 & 22\% & 47\% & 0.9310 & 0.9310 & 0.9167 & 0.9167 & 0.8621 & 0.8966 & 0.8333 & 0.8800 & 1.0000 & 1.0000 & 1.0000 & 1.0000\\
3 & 14\% & 54\% & 1.0000 & 1.0000 & 1.0000 & 1.0000 & 0.9500 & 0.9500 & 0.9565 & 0.9565 & 0.8333 & 0.8889 & 0.8235 & 0.8750\\
4 & 17\% & 14\% & 0.9048 & 0.9048 & 0.7500 & 0.7500 & 1.0000 & 1.0000 & 1.0000 & 1.0000 & 0.9500 & 1.0000 & 0.8571 & 1.0000\\
5 & 28\% & 41\% & 0.9688 & 1.0000 & 0.9630 & 1.0000 & 1.0000 & 1.0000 & 1.0000 & 1.0000 & 0.9286 & 1.0000 & 0.9333 & 1.0000\\\hline
Avg & - & - & 0.9609 & \textbf{0.9672} & 0.9259 & \textbf{0.9333} & 0.9424 & \textbf{0.9593} & 0.9180 & \textbf{0.9451} & 0.9424 & \textbf{0.9778} & 0.9228 & \textbf{0.9750}\\\hline\hline
1 & 50\% & 40\% & 0.9608 & 0.9804 & 0.9474 & 0.9730 & 0.9836 & 0.9672 & 0.9787 & 0.9583 & 0.9649 & 0.9825 & 0.9524 & 0.9756\\
2 & 2\% & 67\% & 1.0000 & 1.0000 & 1.0000 & 1.0000 & 1.0000 & 1.0000 & 1.0000 & 1.0000 & 1.0000 & 1.0000 & 1.0000 & 1.0000\\
3 & 11\% & 42\% & 0.8125 & 0.8125 & 0.7692 & 0.7692 & 1.0000 & 1.0000 & 1.0000 & 1.0000 & 0.8571 & 0.8571 & 0.8333 & 0.8333\\
4 & 6\% & 65\% & 0.8000 & 1.0000 & 0.8000 & 1.0000 & 0.6667 & 0.8333 & 0.5000 & 0.6667 & 0.8571 & 1.0000 & 0.8571 & 1.0000 \\
5 & 32\% & 25\% & 0.9750 & 0.9750 & 0.9565 & 0.9565 & 0.8667 & 0.9000 & 0.7778 & 0.8235 & 0.9714 & 1.0000 & 0.9524 & 1.0000\\\hline
Avg & - & - & 0.9097 & \textbf{0.9536} & 0.8946 & \textbf{0.9397} & 0.9034 & \textbf{0.9401} & 0.8513 & \textbf{0.8897} & 0.9301 & \textbf{0.9679} & 0.9190 & \textbf{0.9618}\\\hline\hline
1 & 52\% & 39\% & 0.9661 & 1.0000 & 0.9545 & 1.0000 & 0.9828 & 0.9828 & 0.9767 & 0.9767 & 0.9821 & 1.0000 & 0.9756 & 1.0000\\
2 & 5\% & 28\% & 1.0000 & 1.0000 & 1.0000 & 1.0000 & 1.0000 & 1.0000 & 1.0000 & 1.0000 & 1.0000 & 1.0000 & 1.0000 & 1.0000\\
3 & 5\% & 38\% & 1.0000 & 1.0000 & 1.0000 & 1.0000 & 0.7500 & 1.0000 & 0.6667 & 1.0000 & 1.0000 & 1.0000 & 1.0000 & 1.0000\\
4 & 14\% & 22\% & 0.9474 & 1.0000 & 0.8889 & 1.0000 & 0.8824 & 0.9412 & 0.5000 & 0.6667 & 0.9412 & 1.0000 & 0.8000 & 1.0000\\
5 & 25\% & 44\% & 0.9259 & 0.8889 & 0.9231 & 0.8889 & 1.0000 & 1.0000 & 1.0000 & 1.0000 & 0.8621 & 0.8276 & 0.8462 & 0.8276\\\hline
Avg & - & - & 0.9679 & \textbf{0.9778} & 0.9533 & \textbf{0.9778} & 0.9230 & \textbf{0.9848} & 0.8287 & \textbf{0.9287} & 0.9571 & \textbf{0.9655} & 0.9244 & \textbf{0.9655}\\\hline
\end{tabular}
\end{table*}

\begin{table*}[ht]
\centering
\tiny
\caption{Experiment 1 (with missing data)}\label{results.breast.imp}
\begin{tabular}{|c| c| c|| c| c|| c| c|| c| c|| c| c|| c| c|| c| c|}
\hline
 \textbf{Par-}& & & \textbf{Accuracy} &         \textbf{Accuracy} & \textbf{F1} & \textbf{F1} & \textbf{Accuracy} &         \textbf{Accuracy} & \textbf{F1} & \textbf{F1} & \textbf{Accuracy} &         \textbf{Accuracy} & \textbf{F1} & \textbf{F1}  \\
 \textbf{tici-} & \textbf{Size} & \textbf{pos (\%)} & \textbf{pre-FL} &         \textbf{post-FL} & \textbf{pre-FL} & \textbf{post-FL} & \textbf{pre Pri-} &         \textbf{post Pri-} & \textbf{pre Pri-} & \textbf{post Pri-} & \textbf{pre encr.} &         \textbf{post encr.} & \textbf{pre encr.} & \textbf{post encr.}  \\
 \textbf{pant} & & & & & & & \textbf{vate FL} & \textbf{vate FL} & \textbf{vate FL} & \textbf{vate FL}  & \textbf{FL} & \textbf{FL} & \textbf{FL} & \textbf{FL} \\
\hline
1 & 20\% & 47\% & 0.9130 & 0.9565 & 0.9167 & 0.9600 & 0.9130 & 1.0000 & 0.8750 & 1.0000 & 0.9565 & 0.9565 & 0.9474 & 0.9474\\
2 & 20\% & 41\% & 0.9130 & 0.9565 & 0.9000 & 0.9524 & 0.9565 & 0.9565 & 0.9474 & 0.9412 & 0.8696 & 0.8696 & 0.8235 & 0.8000 \\
3 & 20\% & 31\% & 0.9565 & 1.0000 & 0.9333 & 1.0000 & 0.9565 & 1.0000 & 0.9474 & 1.0000 & 0.9130 & 0.9565 & 0.8571 & 0.9333\\
4 & 20\% & 35\% & 0.9565 & 1.0000 & 0.9091 & 1.0000 & 0.9130 & 0.8696 & 0.9000 & 0.8571 & 0.9565 & 0.9565 & 0.9333 & 0.9333\\
5 & 20\% & 32\% & 0.9565 & 1.0000 & 0.9091 & 1.0000 & 1.0000 & 1.0000 & 1.0000 & 1.0000 & 1.0000 & 1.0000 & 1.0000 & 1.0000 \\\hline
Avg & - & - & 0.9391 & \textbf{0.9826} & 0.9136 & \textbf{0.9825} & 0.9478 & \textbf{0.9652} & 0.9339 & \textbf{0.9697} & 0.9391 & \textbf{0.9478} & 0.9123 & \textbf{0.9228}\\
\hline\hline
1 & 19\% & 28\% & 0.9583 & 1.0000 & 0.9474 & 1.0000 & 0.8846 & 0.9615 & 0.8696 & 0.9524 & 0.9565 & 1.0000 & 0.9565 & 1.0000\\
2 & 22\% & 47\% & 0.8889 & 0.8889 & 0.8571 & 0.8571 & 1.0000 & 0.9615 & 1.0000 & 0.9565 & 0.9032 & 0.8710 & 0.8696 & 0.8333\\
3 & 14\% & 54\% & 0.9048 & 0.8095 & 0.9231 & 0.8462 & 0.8421 & 0.8947 & 0.8800 & 0.9167 & 1.0000 & 1.0000 & 1.0000 & 1.0000 \\
4 & 17\% & 14\% & 0.9500 & 1.0000 & 0.8571 & 1.0000 & 0.9500 & 0.9500 & 0.8571 & 0.8571 & 0.9474 & 0.9474 & 0.8000 & 0.8000\\
5 & 28\% & 41\% & 0.9583 & 1.0000 & 0.9231 & 1.0000 & 1.0000 & 1.0000 & 1.0000 & 1.0000 & 1.0000 & 1.0000 & 1.0000 & 1.0000 \\\hline
Avg & - & - & 0.9321 & \textbf{0.9397} & 0.9016 & \textbf{0.9407} & 0.9353 & \textbf{0.9536} & 0.9213 & \textbf{0.9365} & 0.9614 & \textbf{0.9637} & 0.9252 & \textbf{0.9267}\\\hline\hline
1 & 50\% & 40\% & 0.9565 & 0.9783 & 0.9545 & 0.9767 & 1.0000 & 1.0000 & 1.0000 & 1.0000 & 0.9783 & 0.9783 & 0.9787 & 0.9787\\
2 & 2\% & 67\% & 1.0000 & 1.0000 & 1.0000 & 1.0000 & 0.6667 & 1.0000 & 0.6667 & 1.0000 & 0.7500 & 1.0000 & 0.6667 & 1.0000 \\
3 & 11\% & 42\% & 0.9375 & 1.0000 & 0.9412 & 1.0000 & 0.9333 & 1.0000 & 0.9474 & 1.0000 & 1.0000 & 0.9333 & 1.0000 & 0.9412\\
4 & 6\% & 65\% & 0.8000 & 0.8000 & 0.8000 & 0.8000 & 1.0000 & 1.0000 & 1.0000 & 1.0000 & 1.0000 & 1.0000 & 1.0000 & 1.0000\\
5 & 32\% & 25\% & 0.9783 & 0.9565 & 0.9474 & 0.8889 & 0.9783 & 0.9565 & 0.9474 & 0.9091 & 0.9783 & 0.9565 & 0.9333 & 0.8571 \\\hline
Avg & - & - & 0.9345 & \textbf{0.9470} & 0.9286 & \textbf{0.9331} & 0.9157 & \textbf{0.9913} & 0.9123 & \textbf{0.9818} & 0.9413 & \textbf{0.9736} & 0.9157 & \textbf{0.9554}\\\hline\hline
1 & 52\% & 39\% & 0.9545 & 0.9773 & 0.9600 & 0.9796 & 0.9524 & 0.9286 & 0.9474 & 0.9189 & 0.9318 & 0.9545 & 0.9362 & 0.9565\\
2 & 5\% & 28\% & 0.8333 & 1.0000 & 0.6667 & 1.0000 & 1.0000 & 1.0000 & 1.0000 & 1.0000 & 1.0000 & 1.0000 & 1.0000 & 1.0000 \\
3 & 5\% & 38\% & 0.8333 & 1.0000 & 0.8000 & 1.0000 & 0.8000 & 1.0000 & 0.8000 & 1.0000 & 1.0000 & 1.0000 & 1.0000 & 1.0000\\
4 & 14\% & 22\% & 0.8824 & 1.0000 & 0.5000 & 1.0000 & 0.8947 & 0.9474 & 0.8000 & 0.8571 & 1.0000 & 1.0000 & 1.0000 & 1.0000\\
5 & 25\% & 44\% & 0.9767 & 0.9302 & 0.9565 & 0.8696 & 1.0000 & 0.9524 & 1.0000 & 0.9286 & 0.9545 & 1.0000 & 0.9167 & 1.0000\\\hline
Avg & - & - & 0.8961 & \textbf{0.9815} & 0.7766 & \textbf{0.9698} & 0.9294 & \textbf{0.9657} & 0.9095 & \textbf{0.9409} & 0.9773 & \textbf{0.9909} & 0.9706 & \textbf{0.9913}\\\hline
\end{tabular}
\end{table*}

\subsection{Experiment 2}


The Chronic Kidney Dataset is the dataset for this experiment. 
It is publicly available at the UC Irvine Machine Learning Repository \cite{dua.2019}.


Tables \ref{results.kidney} and \ref{results.kidney.imp} show the results of the experiments made with this dataset and with the same partitions as the previous experiment.

According to the results, the federation process improves the averaged accuracy and the F1 score over all participants with no missing data for every test. Both the private and non-private approaches result in similar performance metrics, with a small decrease in the case of the balanced datasets, and both increases and decreases for the imbalanced datasets, as in the previous experiment: for the case with no additional privacy layer, for the even split the average accuracy improves from 0.9625 to 0.9750 and the average F1 score from 0.9690 to 0.9787, for the first uneven split the average accuracy goes from 0.9388 to 0.9738 and the average F1 score from 0.9478 to 0.9802, for the second uneven split the average accuracy increases from 0.9331 to 0.9857 and the average F1 score from 0.9493 to 0.9913, and for the last uneven split the average accuracy goes from 0.8506 to 0.9294 and the average F1 score from 0.8793 to 0.9428; and for the case with differential privacy, for the even split the average accuracy improves from 0.9375 to 0.9625 and the average F1 score from 0.9480 to 0.9659, for the first uneven split the average accuracy goes from 0.9473 to 0.9713 and the average F1 score from 0.9509 to 0.9777, for the second uneven split the average accuracy increases from 0.8446 to 0.9192 and the average F1 score from 0.8505 to 0.9400, and for the last uneven split the average accuracy goes from 0.9316 to 0.9770 and the average F1 score from 0.9468 to 0.9830. 

In the case of missing data, we find a similar outcome, with the federation process improving the performance metrics in all cases: in the first case without a layer of differential privacy, for the even split the average accuracy increases from 0.9375 to 0.9875 and the average F1 score from 0.9491 to 0.9913, for the first uneven split the average accuracy goes from 0.9326 to 0.9538 and the average F1 score from 0.9425 to 0.9590, for the second uneven split the average accuracy improves from 0.9533 to 0.9867 and the average F1 score from 0.9578 to 0.9869, and for the third uneven split the average accuracy goes from 0.9179 to 0.9439 and the average F1 score from 0.8971 to 0.9294. In the second case with the layer of differential Privacy, for the even split the average accuracy goes from 0.9375 to 0.9625 and the average F1 score from 0.9473 to 0.9672, for the first uneven split the average accuracy improves from 0.9678 to 0.9895 and the average F1 score from 0.9744 to 0.9913, for the second uneven split the average accuracy increases from 0.9438 to 0.9875 and the average F1 score from 0.9505 to 0.9818, and for the third uneven split the average accuracy goes from 0.8443 to 0.9572 and the average F1 score from 0.8385 to 0.9560. 

Adding an encryption layer using chaotic maps, the results are also positive after the federated learning process both in accuracy and in F1 score. Firstly, if there is no missing data, the accuracy in the even split case goes from 0.95 to 0.9625, and the F1 score from 0.9543 to 0.9682. In the first uneven split, the accuracy increases from 0.9255 to 0.9455, and the F1 score from 0.8961 to 0.9143, in the second uneven split the accuracy improves from 0.9752 to 0.9857 and the F1 score from 0.9799 to 0.9895, and in the last uneven split the accuracy goes from 0.9543 to 0.9907 and the F1 score from 0.9426 to 0.9926. 

When there is missing data and the data is split evenly, the accuracy increases from 0.9625 to 0.9750 and the F1 score from 0.9645 to 0.9750. With the first uneven split the accuracy goes from 0.9689 to 0.9875 and the F1 score from 0.9246 to 0.9920, with the second uneven split the accuracy improves from 0.9204 to 0.9935 and the F1 score from 0.9148 to 0.9959, and in the third uneven split the accuracy increases from 0.9596 to 0.9939 and the F1 score from 0.9543 to 0.9943.

Summarizing these results, in every case the average accuracy and the average F1 score show a performance close to 0.9 in every case except for the first uneven split for the chaotic map encryption, and the third uneven split, where two participants have less than 6\% of the data, that we see that the average accuracy is bigger than 0.85 before the federation process, and bigger than 0.9 after. Also, the Federation Learning process improves the performance metrics in every case, and that the final results are very similar in every experiment, with or without an additional layer of differential privacy or chaotic map encryption, and with or without the imputing of missing data.  

\begin{table*}[ht]
\centering
\tiny
\caption{Experiment 2 (no missing data)}\label{results.kidney}
\begin{tabular}{|c| c| c|| c| c|| c| c|| c| c|| c| c|| c| c|| c| c|}
\hline
 \textbf{Par-}& & & \textbf{Accuracy} &         \textbf{Accuracy} & \textbf{F1} & \textbf{F1} & \textbf{Accuracy} &         \textbf{Accuracy} & \textbf{F1} & \textbf{F1} & \textbf{Accuracy} &         \textbf{Accuracy} & \textbf{F1} & \textbf{F1}  \\
 \textbf{tici-} & \textbf{Size} & \textbf{pos (\%)} & \textbf{pre-FL} &         \textbf{post-FL} & \textbf{pre-FL} & \textbf{post-FL} & \textbf{pre Pri-} &         \textbf{post Pri-} & \textbf{pre Pri-} & \textbf{post Pri-} & \textbf{pre encr.} &         \textbf{post encr.} & \textbf{pre encr.} & \textbf{post encr.}  \\
 \textbf{pant} & & & & & & & \textbf{vate FL} & \textbf{vate FL} & \textbf{vate FL} & \textbf{vate FL}  & \textbf{FL} & \textbf{FL} & \textbf{FL} & \textbf{FL} \\
\hline
1 & 20\% & 73\% & 0.9375 & 0.9375 & 0.9474 & 0.9524 &  0.9375 & 0.9375 & 0.9474 & 0.9474 & 0.8750 & 0.8750 & 0.8889 & 0.9000\\
2 & 20\% & 58\% & 0.9375 & 1.0000 & 0.9412 & 1.0000 &  0.9375 & 0.9375 & 0.9412 & 0.9412 & 0.9375 & 1.0000 & 0.9412 & 1.0000 \\
3 & 20\% & 56\% & 0.9375 & 1.0000 & 0.9565 & 1.0000 &  0.9375 & 1.0000 & 0.9565 & 1.0000 & 1.0000 & 1.0000 & 1.0000 & 1.0000 \\
4 & 20\% & 65\% & 1.0000 & 0.9375 & 1.0000 & 0.9412 &  0.9375 & 0.9375 & 0.9474 & 0.9412 & 0.9375 & 0.9375 & 0.9412 & 0.9412\\
5 & 20\% & 60\% & 1.0000 & 1.0000 & 1.0000 & 1.0000 &  0.9375 & 1.0000 & 0.9474 & 1.0000 & 1.0000 & 1.0000 & 1.0000 & 1.0000\\\hline
Avg & - & - & 0.9625 & \textbf{0.9750} & 0.9690 & \textbf{0.9787} & 0.9375 & \textbf{0.9625} & 0.9480 & \textbf{0.9659} & 0.9500 & \textbf{0.9625} & 0.9543 & \textbf{0.9682}\\
\hline\hline
1 & 14\% & 43\% & 0.9167 & 1.0000 & 0.9091 & 1.0000 & 1.0000 & 1.0000 & 1.0000 & 1.0000 & 1.0000 & 1.0000 & 1.0000 & 1.0000\\
2 & 27\% & 71\% & 0.9552 & 0.9524 & 0.9600 & 0.9600 & 0.8571 & 0.9524 & 0.8800 & 0.9600 & 0.9000 & 1.0000 & 0.9091 & 1.0000\\
3 & 21\% & 76\% & 0.8667 & 1.0000 & 0.9000 & 1.0000 & 1.0000 & 0.9474 & 1.0000 & 0.9630 & 1.0000 & 1.0000 & 1.0000 & 1.0000\\
4 & 14\% & 39\% & 1.0000 & 1.0000 & 1.0000 & 1.0000 & 0.9231 & 1.0000 & 0.9091 & 1.0000 & 0.7273 & 0.7273 & 0.5714 & 0.5714\\
5 & 24\% & 66\% & 0.9583 & 0.9167 & 0.9697 & 0.9412 & 0.9565 & 0.9565 & 0.9655 & 0.9655 & 1.0000 & 1.0000 & 1.0000 & 1.0000 \\\hline
Avg & - & - & 0.9388 & \textbf{0.9738} & 0.9478 & \textbf{0.9802} & 0.9473 & \textbf{0.9713} & 0.9509 & \textbf{0.9777} & 0.9255 & \textbf{0.9455} & 0.8961 & \textbf{0.9143}\\\hline\hline
1 & 49\% & 67\% & 0.9750 & 1.0000 & 0.9804 & 1.0000 & 1.0000 & 1.0000 & 1.0000 & 1.0000 & 1.0000 & 1.0000 & 1.0000 & 1.0000 \\
2 & 4\% & 60\% & 1.0000 & 1.0000 & 1.0000 & 1.0000 & 0.8000 & 1.0000 &  0.8000 & 1.0000 & 1.0000 & 1.0000 & 1.0000 & 1.0000\\
3 & 13\% & 68\% & 0.8571 & 0.9286 & 0.9091 & 0.9565 & 0.9231 & 0.8462 & 0.9524 & 0.9000 & 0.9286 & 0.9286 & 0.9474 & 0.9474\\
4 & 3\% & 63\% & 0.8333 & 1.0000 & 0.8571 & 1.0000 & 0.5000 & 0.7500 & 0.5000 & 0.8000 & 1.0000 & 1.0000 & 1.0000 & 1.0000\\
5 & 30\% & 54\% & 1.0000 & 1.0000 & 1.0000 & 1.0000 & 1.0000 & 1.0000 & 1.0000 &  1.0000 & 0.9474 & 1.0000 & 0.9524 & 1.0000\\\hline
Avg & - & - & 0.9331 & \textbf{0.9857} & 0.9493 & \textbf{0.9913} & 0.8446 & \textbf{0.9192} & 0.8505 & \textbf{0.9400} & 0.9752 & \textbf{0.9857} & 0.9799 & \textbf{0.9895}\\\hline\hline
1 & 52\% & 66\% & 1.0000 & 1.0000 & 1.0000 & 1.0000 & 0.9535 & 0.9302 & 0.9643 & 0.9454 & 0.9535 & 0.9535 & 0.9630 & 0.9630\\
2 & 5\% & 33\% & 1.0000 & 1.0000 & 1.0000 & 1.0000 & 1.0000 & 1.0000 & 1.0000 & 1.0000 & 1.0000 & 1.0000 & 1.0000 & 1.0000\\
3 & 3\% & 29\% & 0.6000 & 0.8000 & 0.7500 & 0.8571 & 0.7500 & 1.0000 & 0.8000 & 1.0000 & 1.0000 & 1.0000 & 1.0000 & 1.0000\\
4 & 15\% & 34\% & 0.7778 & 0.8889 & 0.7500 & 0.8889 & 1.0000 & 1.0000 & 1.0000 & 1.0000 & 0.8182 & 1.0000 & 0.7500 & 1.0000\\
5 & 25\% & 82\% & 0.8750 & 0.9583 & 0.8966 & 0.9677 & 0.9545 & 0.9545 & 0.9697 & 0.9697 & 1.0000 & 1.0000 & 1.0000 & 1.0000\\\hline
Avg & - & - & 0.8506 & \textbf{0.9294} & 0.8793 & \textbf{0.9428} & 0.9316 & \textbf{0.9770} & 0.9468 & \textbf{0.9830} & 0.9543 & \textbf{0.9907} & 0.9426 & \textbf{0.9926}\\\hline
\end{tabular}
\end{table*}

\begin{table*}[ht]
\centering
\tiny
\caption{Experiment 2 (with missing data)}\label{results.kidney.imp}
\begin{tabular}{|c| c| c|| c| c|| c| c|| c| c|| c| c|| c| c|| c| c|}
\hline
 \textbf{Par-}& & & \textbf{Accuracy} &         \textbf{Accuracy} & \textbf{F1} & \textbf{F1} & \textbf{Accuracy} &         \textbf{Accuracy} & \textbf{F1} & \textbf{F1} & \textbf{Accuracy} &         \textbf{Accuracy} & \textbf{F1} & \textbf{F1}  \\
 \textbf{tici-} & \textbf{Size} & \textbf{pos (\%)} & \textbf{pre-FL} &         \textbf{post-FL} & \textbf{pre-FL} & \textbf{post-FL} & \textbf{pre Pri-} &         \textbf{post Pri-} & \textbf{pre Pri-} & \textbf{post Pri-} & \textbf{pre encr.} &         \textbf{post encr.} & \textbf{pre encr.} & \textbf{post encr.}  \\
 \textbf{pant} & & & & & & & \textbf{vate FL} & \textbf{vate FL} & \textbf{vate FL} & \textbf{vate FL}  & \textbf{FL} & \textbf{FL} & \textbf{FL} & \textbf{FL} \\
\hline
1 & 20\% & 73\% & 0.8750 & 1.0000 & 0.8889 & 1.0000 &  1.0000 & 1.0000 & 1.0000 & 1.0000 & 1.0000 & 1.0000 & 1.0000 & 1.0000 \\
2 & 20\% & 58\% & 0.8750 & 0.9375 & 0.9091 & 0.9565 &  1.0000 & 0.9375 & 1.0000 & 0.9474 & 0.8750 & 0.8750 & 0.8750 & 0.8750 \\
3 & 20\% & 56\% & 1.0000 & 1.0000 & 1.0000 & 1.0000 &  0.8750 & 0.9375 & 0.9000 & 0.9412 & 1.0000 & 1.0000 & 1.0000 & 1.0000 \\
4 & 20\% & 65\% & 0.9375 & 1.0000 & 0.9474 & 1.0000 &  0.9375 & 0.9375 & 0.9474 & 0.9474 & 1.0000 & 1.0000 & 1.0000 & 1.0000\\
5 & 20\% & 60\% & 1.0000 & 1.0000 & 1.0000 & 1.0000 &  0.8750 & 1.0000 & 0.8889 & 1.0000 & 0.9375 & 1.0000 & 0.9474 & 1.0000\\\hline
Avg & - & - & 0.9375 & \textbf{0.9875} & 0.9491 & \textbf{0.9913} & 0.9375 & \textbf{0.9625} & 0.9473 & \textbf{0.9672} & 0.9625 & \textbf{0.9750} & 0.9645 & \textbf{0.9750}\\
\hline\hline
1 & 14\% & 43\% & 1.0000 & 1.0000 & 1.0000 & 1.0000 & 1.0000 & 1.0000 & 1.0000 & 1.0000 & 0.9444 & 1.0000 & 0.9565 & 1.0000\\
2 & 27\% & 71\% & 0.9130 & 0.9565 & 0.9286 & 0.9677 & 0.8947 & 0.9474 & 0.9091 & 0.9565 & 1.0000 & 1.0000 & 1.0000 & 1.0000\\
3 & 21\% & 76\% & 0.8750 & 0.9375 & 0.9091 & 0.9524 & 0.9444 & 1.0000 & 0.9630 & 1.0000 & 1.0000 & 0.9375 & 1.0000 & 0.9600\\
4 & 14\% & 39\% & 1.0000 & 1.0000 & 1.0000 & 1.0000 & 1.0000 & 1.0000 & 1.0000 & 1.0000 & 0.9000 & 1.0000 & 0.6667 & 1.0000\\
5 & 24\% & 66\% & 0.8750 & 0.8750 & 0.8750 & 0.8750 & 1.0000 & 1.0000 & 1.0000  & 1.0000 & 1.0000 & 1.0000 & 1.0000 & 1.0000 \\\hline
Avg & - & - & 0.9326 & \textbf{0.9538} & 0.9425 & \textbf{0.9590} & 0.9678 & \textbf{0.9895} & 0.9744 & \textbf{0.9913} & 0.9689 & \textbf{0.9875} & 0.9246 & \textbf{0.9920}\\
\hline\hline
1 & 49\% & 67\% & 0.86667 & 0.9667 & 0.9091  & 0.9778 & 1.0000 & 1.0000 & 1.0000 & 1.0000 & 0.9355 & 0.9677 & 0.9600 & 0.9796\\
2 & 4\% & 60\% & 1.0000 & 1.0000 & 1.0000  & 1.0000 & 1.0000 & 1.0000 & 1.0000 & 1.0000 & 0.7500 & 1.0000 & 0.6667 & 1.0000\\
3 & 13\% & 68\% & 1.0000 & 1.0000 & 1.0000  & 1.0000 & 1.0000 & 1.0000 & 1.0000 & 1.0000 & 0.9167 & 1.0000 & 0.9474 & 1.0000\\
4 & 3\% & 63\% & 1.0000 & 1.0000 & 1.0000 & 1.0000 & 0.7500 & 1.0000 & 0.8000 & 1.0000 & 1.0000 & 1.0000 & 1.0000 & 1.0000\\
5 & 30\% & 54\% & 0.9000 & 0.9667 & 0.8800 & 0.9565 & 0.9688 & 0.9375 & 0.9524 & 0.9091 & 1.0000 & 1.0000 & 1.0000 & 1.0000\\\hline
Avg & - & - & 0.9533 & \textbf{0.9867} & 0.9578 & \textbf{0.9869} & 0.9438 & \textbf{0.9875} & 0.9505 & \textbf{0.9818} & 0.9204 & \textbf{0.9935} & 0.9148 & \textbf{0.9959}\\
\hline\hline
1 & 52\% & 66\% & 0.9394 & 0.9697 & 0.9615 & 0.9804 & 0.9355  & 0.9677  & 0.9583 & 0.9796 & 1.0000 & 1.0000 & 1.0000 & 1.0000\\
2 & 5\% & 33\% & 0.7500 & 0.7500 & 0.6667 & 0.6667  & 0.7500  & 1.0000  & 0.6667 & 1.0000 & 1.0000 & 1.0000 & 1.0000 & 1.0000\\
3 & 3\% & 29\% & 1.0000 & 1.0000 & 1.0000 & 1.0000  & 0.7500  & 1.0000  & 0.8000 & 1.0000 & 1.0000 & 1.0000 & 1.0000 & 1.0000\\
4 & 15\% & 34\% & 0.9000 & 1.0000 & 0.8571 & 1.0000 & 0.8182  & 0.8182  & 0.8000 & 0.8000 & 0.8889 & 1.0000 & 0.8571 & 1.0000\\
5 & 25\% & 82\% & 1.0000 & 1.0000 & 1.0000 & 1.0000 & 0.9677  & 1.0000  & 0.9677 & 1.0000 & 0.9091 & 0.9697 & 0.9143 & 0.9714\\\hline
Avg & - & - & 0.9179 & \textbf{0.9439} & 0.8971 & \textbf{0.9294} & 0.8443 & \textbf{0.9572} & 0.8385 & \textbf{0.9560} & 0.9596 & \textbf{0.9939} & 0.9543 & \textbf{0.9943}\\
\hline
\end{tabular}
\end{table*}

\subsection{Experiment 3}



The Parkinson's dataset is the dataset for the third experiment. 
It is publicly available at the UC Irvine Machine Learning Repository \cite{dua.2019}.

The results for the experiments with the Parkinson's disease dataset are shown in Tables \ref{results.parkinson} and \ref{results.parkinson.imp}. As in the previous experiments, the averaged accuracy and F1 score of the models are improved after the Federated Learning procedure in every case, and both the traditional, the differential privacy, and the chaotic map approaches, and the experiments with or without missing data, reach similar performance metrics. 

As a summary of the results: when dealing without missing data and the experiments without the additional layer of differential privacy, the average accuracy goes from 0.6893 to 0.7929 in the even split, from 0.8288 to 0.8429 in the first uneven split, from 0.6065 to 0.8104 in the second uneven split, and from 0.7714 to 0.8254 in the last uneven split, and the average F1 score increases from 0.7738 to 0.8598 for the even split, from 0.8894 to 0.9049 for the first uneven split, from 0.6495 to 0.8678 for the second uneven split, and from 0.8345 to 0.8646 for the third uneven split. With the differential privacy layer, the average accuracy improves from 0.6643 to 0.7964, from 0.7429 to 0.8596, from 0.7389 to 0.9500, and from 0.8181 to 0.9219, and the average F1 score increases from 0.7506 to 0.8444, from 0.7961 to 0.8958, from 0.8149 to 0.9636, and from 0.8701 to 0.9492 for the even split, the first, the second and the third uneven split respectively. Including the chaotic map encryption, the accuracy improves from 0.6964 to 0.7464 and the F1 score from 0.7957 to 0.8078 in the case of the even split dataset, with the first uneven split the accuracy goes from 0.8267 to 0.9022 and the F1 score from 0.8814 to 0.9330, for the second uneven split the accuracy increases from 0.7793 to 0.8079 and the F1 score from 0.7160 to 0.7338, and for the third case, the accuracy goes from 0.8556 to 0.8667 and the F1 score from 0.9 to 0.9091.

When imputing missing data, when there is no additional privacy layer the average accuracy increases from 0.7250 to 0.7750, from 0.7405 to 0.7690, from 0.8122 to 0.8344, and from 0.8271 to 0.8857, and the average F1 score goes from 0.8220 to 0.8513, from 0.8133 to 0.8352, from 0.8631 to 0.8898, and from 0.8616 to 0.9095 for the even split, the first, the second and the third uneven split respectively. With the additional differential privacy layer the performance metrics improve from 0.7500 to 0.8000 for the average accuracy and from 0.8248 to 0.8609 for the average F1 score for the even split, from 0.7419 to 0.9262 for the average accuracy and from 0.8305 to 0.9559 for the average F1 score for the first uneven split, from 0.8159 to 0.9270 for the average accuracy and from 0.8500 to 0.9548 for the average F1 score for the second, and from 0.6886 to 0.8076 for the average accuracy and from 0.6933 to 0.8574 for the average F1 score for the third. Finally, when adding the chaotic map encryption, the accuracy improves from 0.75, 0.8133, 0.7044, and 0.6933 to 0.8750, 0.8483, 0.8467, and 0.72 in the even split, and first, second and third uneven split respectively, and the F1 score goes from 0.8238, 0.8425, 0.7814, and 0.6698 to 0.9198, 0.9006, 0.8987, and 0.685 in the same cases.

Again we can see that in every case the Federated Learning process improves the performance metrics, not only in the traditional case but also in the experiments when we improve the security of the system by adding the additional layer of differential privacy or the chaotic map encryption, and when we impute missing data for one of the participants. 

\begin{table*}[ht]
\centering
\tiny
\caption{Experiment 3 (no missing data)}\label{results.parkinson}
\begin{tabular}{|c| c| c|| c| c|| c| c|| c| c|| c| c|| c| c|| c| c|}
\hline
 \textbf{Par-}& & & \textbf{Accuracy} &         \textbf{Accuracy} & \textbf{F1} & \textbf{F1} & \textbf{Accuracy} &         \textbf{Accuracy} & \textbf{F1} & \textbf{F1} & \textbf{Accuracy} &         \textbf{Accuracy} & \textbf{F1} & \textbf{F1}  \\
 \textbf{tici-} & \textbf{Size} & \textbf{pos (\%)} & \textbf{pre-FL} &         \textbf{post-FL} & \textbf{pre-FL} & \textbf{post-FL} & \textbf{pre Pri-} &         \textbf{post Pri-} & \textbf{pre Pri-} & \textbf{post Pri-} & \textbf{pre encr.} &         \textbf{post encr.} & \textbf{pre encr.} & \textbf{post encr.}  \\
 \textbf{pant} & & & & & & & \textbf{vate FL} & \textbf{vate FL} & \textbf{vate FL} & \textbf{vate FL}  & \textbf{FL} & \textbf{FL} & \textbf{FL} & \textbf{FL} \\
\hline
1 & 20\% & 65\% & 0.5714 & 0.7143 & 0.6667 & 0.8000 &  0.5714 & 0.8571 & 0.6667 & 0.9091 & 0.8571 & 0.8571 & 0.9091 & 0.8889\\
2 & 20\% & 78\% & 1.0000 & 1.0000 & 1.0000 & 1.0000 &  0.8750 & 0.8750 & 0.9231 & 0.9231 & 0.7500 & 0.8750 & 0.8333 & 0.9231\\
3 & 20\% & 75\% & 0.5000 & 0.8750 & 0.6000 & 0.9091 &  0.6250 & 0.6250 & 0.6667 & 0.6667 & 0.7500 & 0.6250 & 0.8000 & 0.6667\\
4 & 20\% & 78\% & 0.7500 & 0.500 & 0.8333 & 0.6667 &  0.6250 & 0.7500 & 0.7692 & 0.8000 & 0.6250 & 0.6250 & 0.7692 & 0.7273\\
5 & 20\% & 79\% & 0.6250 & 0.8750 & 0.7692 & 0.9231 &  0.6250 & 0.8750 & 0.7273 & 0.9231 & 0.5000 & 0.7500 & 0.6667 & 0.83333 \\\hline
Avg & - & - & 0.6893 & \textbf{0.7929} & 0.7738 & \textbf{0.8598} & 0.6643 & \textbf{0.7964} & 0.7506 & \textbf{0.8444} & 0.6964 & \textbf{0.7464} & 0.7957 & \textbf{0.8078}\\
\hline\hline
1 & 14\% & 69\% & 0.6667 & 0.6667 & 0.8000 & 0.8000 & 0.2500 & 0.7500 & 0.4000 & 0.8000 & 0.6667 & 0.8333 & 0.7500 & 0.8571\\
2 & 24\% & 75\% & 0.7500 & 0.7500 & 0.8235 & 0.8421 & 0.8889 & 0.8889 & 0.9333 & 0.9333 & 1.0000 & 1.0000 & 1.0000 & 1.0000\\
3 & 19\% & 86\% & 1.0000 & 0.8889 & 1.0000 & 0.9412 & 0.9091 & 0.9091 & 0.9474 & 0.9474 & 0.6667 & 0.7778 & 0.8000 & 0.8750\\
4 & 10\% & 50\% & 1.0000 & 1.0000 & 1.0000 & 1.0000 & 0.8333 & 0.8333 & 0.8000 & 0.8571 & 1.0000 & 1.0000 & 1.0000 & 1.0000\\
5 & 32\% & 78\% & 0.7273 & 0.9091 & 0.8235 & 0.9412 & 0.8333 & 0.9167 & 0.9000  & 0.9412 & 0.8000 & 0.9000 & 0.8571 & 0.9333\\\hline
Avg & - & - & 0.8288 & \textbf{0.8429} & 0.8894 & \textbf{0.9049} & 0.7429 & \textbf{0.8596} & 0.7961 & \textbf{0.8958} & 0.8267 & \textbf{0.9022} & 0.8814 & \textbf{0.9330}\\
\hline\hline
1 & 21\% & 75\% & 0.8182 & 0.9091 & 0.8750 & 0.9412 & 0.8889 & 1.0000 & 0.9231 & 1.0000 & 1.0000 & 1.0000 & 1.0000 & 1.0000\\
2 & 41\% & 79\% & 0.6429 & 0.7857 & 0.7059 & 0.8421 & 0.7500 & 0.7500 &  0.8182 & 0.8182 & 0.8824 & 0.8824 & 0.9231 & 0.9231\\
3 & 16\% & 83\% & 1.0000 & 1.0000 & 1.0000 & 1.0000 & 1.0000 & 1.0000 & 1.0000 & 1.0000 & 0.8000 & 0.8000 & 0.8571 & 0.8571\\
4 &  8\% & 67\% & 0.0000 & 0.5000 & 0.0000 & 0.6667 & 0.5000 & 1.0000 & 0.6667 & 1.0000 & 0.5000 & 0.5000 & 0.0000 & 0.0000\\
5 & 16\% & 67\% & 0.5714 & 0.8571 & 0.6667 & 0.8889 & 0.5556 & 1.0000 & 0.6667 & 1.0000 & 0.7143 & 0.8571 & 0.8000 & 0.8889\\\hline
Avg & - & - & 0.6065 & \textbf{0.8104} & 0.6495 & \textbf{0.8678} & 0.7389 & \textbf{0.9500} & 0.8149 & \textbf{0.9636} & 0.7793 & \textbf{0.8079} & 0.7160 & \textbf{0.7338}\\
\hline\hline
1 & 55\% & 81\% & 0.8571 & 0.9048 & 0.9032 & 0.9333 & 0.8571 & 0.8095 & 0.9032 & 0.8571 & 1.0000 & 0.9444 & 1.0000 & 0.9655\\
2 & 10\% & 75\% & 1.0000 & 1.0000 & 1.0000 & 1.0000 & 1.0000 & 1.0000 & 1.0000 & 1.0000 & 0.5000 & 0.7500 & 0.6667 & 0.8571\\
3 & 8\% & 67\% & 0.3333 & 0.6667 & 0.5000 & 0.8000 & 0.3333 & 1.0000 & 0.5000 & 1.0000 & 1.0000 & 1.0000 & 1.0000 & 1.0000\\
4 & 9\% & 50\% & 1.0000 & 0.6667 & 1.0000 & 0.6667 & 1.0000 & 1.0000 & 1.0000 & 1.0000 & 1.0000 & 0.7500 & 1.0000 & 0.8000\\
5 & 18\% & 76\% & 0.6667 & 0.8889 & 0.7692 & 0.9231 & 0.9000 & 0.8000 & 0.9474 & 0.8889 & 0.7778 & 0.8889 & 0.8333 & 0.9231\\\hline
Avg & - & - & 0.7714 & \textbf{0.8254} & 0.8345 & \textbf{0.8646} & 0.8181 & \textbf{0.9219} & 0.8701 & \textbf{0.9492} & 0.8556 & \textbf{0.8667} & 0.9000 & \textbf{0.9091}\\
\hline
\end{tabular}
\end{table*}

\begin{table*}[ht]
\centering
\tiny
\caption{Experiment 3 (with missing data)}\label{results.parkinson.imp}
\begin{tabular}{|c| c| c|| c| c|| c| c|| c| c|| c| c|| c| c|| c| c|}
\hline
 \textbf{Par-}& & & \textbf{Accuracy} &         \textbf{Accuracy} & \textbf{F1} & \textbf{F1} & \textbf{Accuracy} &         \textbf{Accuracy} & \textbf{F1} & \textbf{F1} & \textbf{Accuracy} &         \textbf{Accuracy} & \textbf{F1} & \textbf{F1}  \\
 \textbf{tici-} & \textbf{Size} & \textbf{pos (\%)} & \textbf{pre-FL} &         \textbf{post-FL} & \textbf{pre-FL} & \textbf{post-FL} & \textbf{pre Pri-} &         \textbf{post Pri-} & \textbf{pre Pri-} & \textbf{post Pri-} & \textbf{pre encr.} &         \textbf{post encr.} & \textbf{pre encr.} & \textbf{post encr.}  \\
 \textbf{pant} & & & & & & & \textbf{vate FL} & \textbf{vate FL} & \textbf{vate FL} & \textbf{vate FL}  & \textbf{FL} & \textbf{FL} & \textbf{FL} & \textbf{FL} \\
\hline
1 & 20\% & 65\% & 0.6250 & 0.7500 & 0.7692 & 0.8333 & 0.7500 & 0.7500 & 0.8333 & 0.8000 & 0.7500 & 0.8750 & 0.8333 & 0.9091\\
2 & 20\% & 78\% & 0.7500 & 0.7500 & 0.8571 & 0.8333 & 0.7500 & 0.6250 & 0.8571 & 0.7692 & 0.6250 & 0.7500 & 0.7692 & 0.8333\\
3 & 20\% & 75\% & 0.8750 & 0.7500 & 0.9231 & 0.8333 & 0.7500 & 0.8750 & 0.8333 & 0.9231 & 0.7500 & 0.8750 & 0.8333 & 0.9231\\
4 & 20\% & 78\% & 0.6250 & 0.7500 & 0.7273 & 0.8333 & 0.5000 & 0.8750 & 0.6000 & 0.8889 & 0.8750 & 0.8750 & 0.9333 & 0.9333\\
5 & 20\% & 79\% & 0.7500 & 0.8750 & 0.8333 & 0.9231 & 1.0000 & 0.8750 & 1.0000 & 0.9231 & 0.7500 & 1.0000 & 0.7500 & 1.0000\\\hline
Avg & - & - & 0.7250 & \textbf{0.7750} & 0.8220 & \textbf{0.8513} & 0.7500 & \textbf{0.8000} & 0.8248 & \textbf{0.8609} & 0.7500 & \textbf{0.8750} & 0.8238 & \textbf{0.9198}\\
\hline\hline
1 & 14\% & 69\% & 0.7143 & 0.8571 & 0.8000 & 0.9091 & 0.4286 & 0.8571 & 0.6000 & 0.9231 & 0.7500 & 0.7500 & 0.8571 & 0.8571\\
2 & 24\% & 75\% & 0.6667 & 0.6667 & 0.8000 & 0.8000 & 0.6667 & 0.9167 & 0.7778 & 0.9474 & 0.9167 & 0.9167 & 0.9412 & 0.9333\\
3 & 19\% & 86\% & 1.0000 & 1.0000 & 1.0000 & 1.0000 & 0.9000 & 1.0000 & 0.9412 & 1.0000 & 0.9000 & 0.7000 & 0.9474 & 0.8235\\
4 & 10\% & 50\% & 0.7500 & 0.7500 & 0.8000 & 0.8000 & 1.0000 & 1.0000 & 1.0000 & 1.0000 & 0.7500 & 1.0000 & 0.6667 & 1.0000\\
5 & 32\% & 78\% & 0.5714 & 0.5714 & 0.6667 & 0.6667 & 0.7143 & 0.8571 & 0.8333 & 0.9091 & 0.7500 & 0.8750 & 0.8000 & 0.8889\\\hline
Avg & - & - & 0.7405 & \textbf{0.7690} & 0.8133 & \textbf{0.8352} & 0.7419 & \textbf{0.9262} & 0.8305 & \textbf{0.9559} & 0.8133 & \textbf{0.8483} & 0.8425 & \textbf{0.9006}\\
\hline\hline
1 & 21\% & 75\% & 0.7778 & 0.7778 & 0.8750 & 0.8750 & 1.0000 & 1.0000 & 1.0000 & 1.0000 & 0.7778 & 0.7778 & 0.8333 & 0.8571\\
2 & 41\% & 79\% & 0.8667 & 0.8667 & 0.9167 & 0.9167 & 0.8571 & 0.8571 & 0.9167 & 0.9167 & 0.8000 & 0.7333 & 0.8571 & 0.8182\\
3 & 16\% & 83\% & 0.7500 & 0.7500 & 0.8571 & 0.8571 & 1.0000 & 1.0000 & 10000. & 1.0000 & 0.6667 & 0.8333 & 0.7500 & 0.9091\\
4 & 8\% & 67\%  & 1.0000 & 1.0000 & 1.0000 & 1.0000 & 0.6667 & 1.0000 & 0.6667 & 1.0000 & 0.5000 & 1.0000 & 0.6667 & 1.0000\\
5 & 16\% & 67\% & 0.6667 & 0.7778 & 0.6667 & 0.8000 & 0.5556 & 0.7778 & 0.6667 & 0.8571 & 0.7778 & 0.8889 & 0.8000 & 0.9091\\\hline
Avg & - & - & 0.8122 & \textbf{0.8344} & 0.8631 & \textbf{0.8898} & 0.8159 & \textbf{0.9270} & 0.8500 & \textbf{0.9548} & 0.7044 & \textbf{0.8467} & 0.7814 & \textbf{0.8987}\\
\hline\hline
1 & 55\% & 81\% & 0.9286 & 1.0000 & 0.9524 & 1.0000 & 1.0000 & 0.9286 & 1.0000 & 0.9630 & 1.0000 & 1.0000 & 1.0000 & 1.0000\\
2 & 10\% & 75\% & 1.0000 & 0.7500 & 1.0000 & 0.8000 & 1.0000 & 1.0000 & 1.0000 & 1.0000 & 0.4000 & 0.8000 & 0.5714 & 0.8889\\
3 & 8\% & 67\%  & 0.6000 & 1.0000 & 0.6667 & 1.0000 & 0.8000 & 0.8000 & 0.8000 & 0.8571 & 1.0000 & 0.6667 & 1.0000 & 0.6667\\
4 & 9\% & 50\%  & 0.7500 & 0.7500 & 0.8000 & 0.8000 & 0.0000 & 0.6667 & 0.0000 & 0.8000 & 0.3333 & 0.3333 & 0.0000 & 0.0000\\
5 & 18\% & 76\% & 0.8571 & 0.9286 & 0.8889 & 0.9474 & 0.6429 & 0.6429 & 0.6667 & 0.6667 & 0.7333 & 0.8000 & 0.7778 & 0.8696\\\hline
Avg & - & - & 0.8271 & \textbf{0.8857} & 0.8616 & \textbf{0.9095} & 0.6886 & \textbf{0.8076} & 0.6933 & \textbf{0.8574} & 0.6933 & \textbf{0.7200} & 0.6698 & \textbf{0.6850}\\
\hline
\end{tabular}
\end{table*}

\section{Conclusions}\label{conclusions}

The different experiments show that, even in the most imbalanced cases, the Federated Learning process improves the average metrics of the models, increasing their performance, both in accuracy and F1 score, and for both the private, non-private and encrypted approaches. We can also conclude that using an additional layer of encryption and ensuring the privacy of the process does not affect the performance metrics of the model, when compared with the non-private Federated Learning.

%
%

This research proofs that the averaged accuracy and F1 score improves not only in the case where every participants has the same amount of data, but also in cases where there are sharp differences between the volume of data that the participants have. This way, this manuscript includes the hypothetical case where several participants want to train an accurate deep learning model and share it among them, even although some of those participants have much less available data. 

In every experiment, an additional layer of differential privacy and chaotic map encryption was added to ensure the privacy and encryption of the data and compared with the Federated Learning approach without this layer, finding that the performance results of both models are extremely similar. 

Moreover, in order to simulate real cases, the authors also test the same experiments in the event that one of the participants has a missing feature. As in the previous experiments, the performance of the models is improved by the federation process in all cases, for the accuracy and F1 metric and the private, encrypted and non-private cases. 

This approach could be adopted to improve the models used to diagnose diseases, such as breast cancer, chronic kidney disease, Parkinson's, or potentially anyone else, as this paper have shown in the experiments.

\bibliographystyle{splncs04}
\bibliography{main}

\begin{IEEEbiography}[{\includegraphics[width=1in,height=1.25in,clip,keepaspectratio]{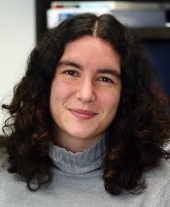}}]{Irina Ar\'evalo} is a Data Scientist and researcher in Artificial Intelligence. She holds a PhD in Mathematics, and is currently a PhD candidate at the University Pablo de Olavide (Seville, Spain). Irina Ar\'evalo also has several years of experience as a Data Scientist in several fields of expertise, including Finance, Consultancy, and Communications. Her current research interests include Distributed Artificial Intelligence, Explainable Artificial Intelligence, and Bias \& Fairness.
\end{IEEEbiography}

\begin{IEEEbiography}[{\includegraphics[width=1in,height=1.25in,clip,keepaspectratio]{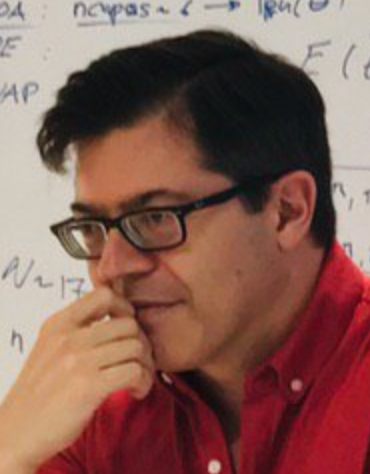}}]{Jose L. Salmeron} is a Professor of Artificial Intelligence at CUNEF University and an AI Senior Research Associate at the University Aut\'onoma of Chile (Chile). His research papers have been published in prestigious journals such as IEEE Transactions on Cybernetics, IEEE Transactions on Fuzzy Systems, IEEE Transactions on Software Engineering, Expert Systems with Applications, Communications of the ACM, Journal of Systems and Software, Computer Standards \& Interfaces, Applied Soft Computing, Engineering Applications of Artificial Intelligence, Neurocomputing, and Information Sciences, among others. Additionally, he is recognized as an ACM lifetime senior member. Currently, his research interests encompass privacy-preserving computing, Distributed Artificial Intelligence, explainable artificial intelligence, and quantum machine learning.
\end{IEEEbiography}
\end{document}